\documentclass[11pt,a4paper]{article}
\usepackage[hyperref]{acl2020}
\usepackage{times}
\usepackage{latexsym}

\usepackage[para]{footmisc}
\usepackage{footnote}
\usepackage{microtype}
\usepackage{booktabs}
\usepackage{listings}
\usepackage{amsmath}
\usepackage{graphicx} 
\usepackage{tabularx}

\usepackage{color, soul}
\usepackage{makecell}

\aclfinalcopy % Uncomment this line for the final submission
\def\aclpaperid{2171} %  Enter the acl Paper ID here

\newcommand{\squishlist}{ 
   \begin{list}{$\bullet$}
    { \setlength{\itemsep}{0pt}      \setlength{\parsep}{2pt} 
      \setlength{\topsep}{3pt}       \setlength{\partopsep}{0pt}
      \setlength{\leftmargin}{1em} \setlength{\labelwidth}{1em}
      \setlength{\labelsep}{0.5em} } }

\newcommand{\squishend}{
    \end{list}  } 
    
\title{Smart To-Do : Automatic Generation of To-Do Items from Emails}

\author{Sudipto Mukherjee$^\dag$\thanks{Work done as an intern at Microsoft Research.\vspace{0.1em}} \qquad Subhabrata Mukherjee$^\ddag$ \qquad Marcello Hasegawa$^\ddag$ 
\\ \textbf{Ahmed Hassan Awadallah}$^\ddag$ \qquad \textbf{Ryen White}$^\ddag$
\\
  $^\dag$University of Washington, Seattle  \qquad$^\ddag$Microsoft Research AI  \\ 
  \texttt{sudipm@uw.edu},\\ \texttt{\{submukhe, marcellh, hassanam, ryenw\}@microsoft.com}}

\date{}

\begin{document}
\maketitle
\begin{abstract}
Intelligent features in email service applications aim to increase productivity by helping people organize their folders, compose their emails and respond to pending tasks. In this work, we explore a new application, Smart-To-Do, that helps users with task management over emails. We introduce a new task and dataset for automatically generating To-Do items from emails where the sender has promised to perform an action. We design a two-stage process leveraging recent advances in neural text generation and sequence-to-sequence learning, obtaining BLEU and ROUGE scores of $0.23$ and $0.63$ for this task. To the best of our knowledge, this is the first work to address the problem of composing To-Do items from emails. 
\end{abstract}

\section{Introduction}

Email is one of the most used forms of communication especially in enterprise and work settings \citep{radicati2015email}. With the growing number of users in email platforms, service providers are constantly seeking to improve user experience for a myriad of applications such as online retail, instant messaging and events management \citep{googleCalendar}. Smart Reply \cite{kannan2016smart} and Smart Compose \citep{chen2019smart} are two recent features that provide contextual assistance to users aiming to reduce typing efforts. Another line of work in this direction is for automated task management and scheduling.  %While replying to emails is a quintessential task, the evolving email platforms also aim to help users manage their tasks. 
For example. the recent Nudge feature\footnote{\href{https://www.makeuseof.com/tag/what-is-nudge-gmail/}{Gmail Nudge}} in Gmail and Insights in Outlook\footnote{ \href{https://docs.microsoft.com/en-us/workplace-analytics/myanalytics/use/add-in}{Outlook Insights}} are designed to remind users to follow-up on an email or pay attention to pending tasks.

%pose the question : \textit{Can we generate an automatic To-Do List for the user based on email replies promising actions ?}

\begin{figure}[t]
\includegraphics[width=0.47\textwidth]{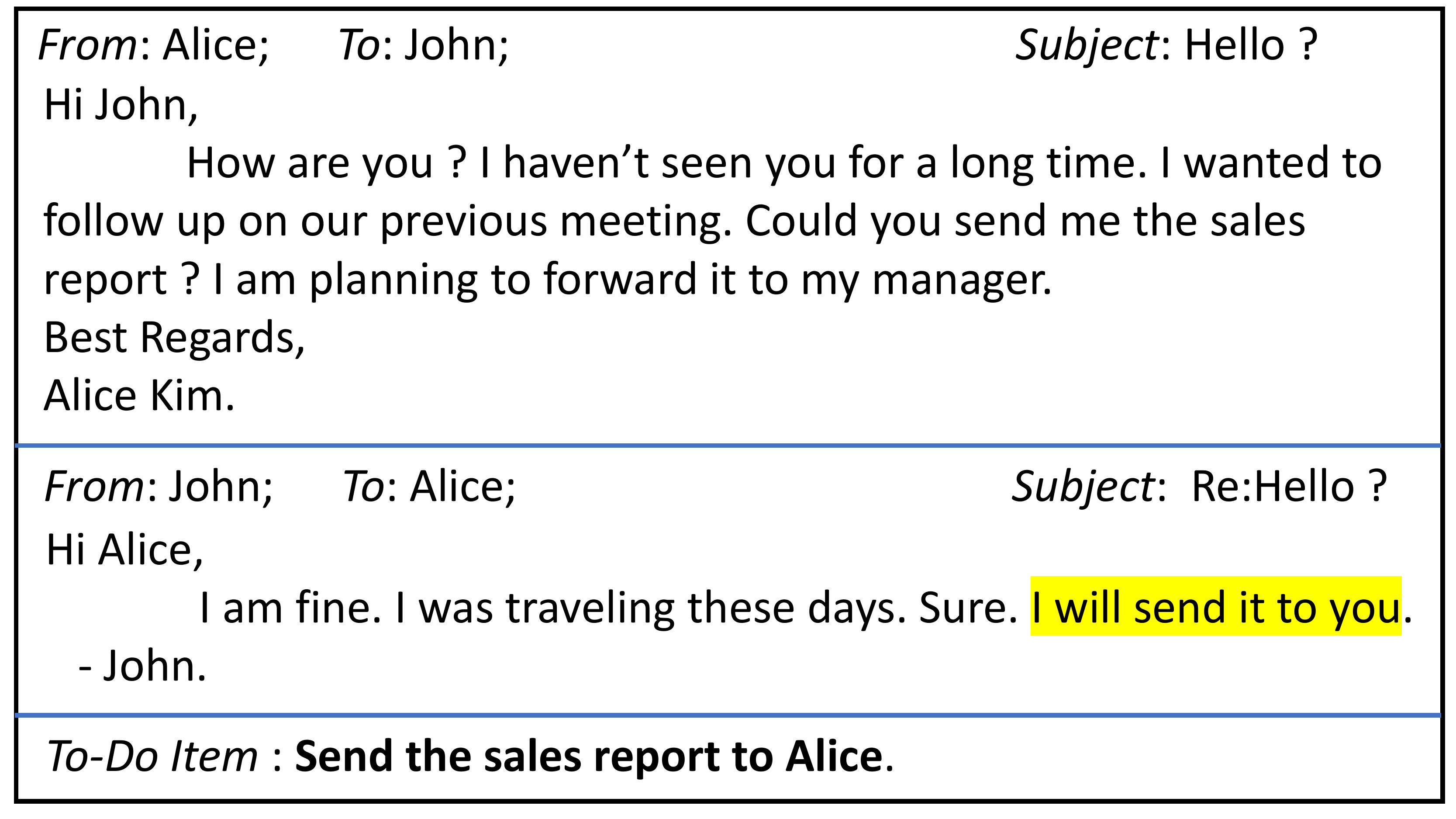} 
\vspace{-1em}
\caption{An illustration showing the email and a commitment sentence (in yellow) and the target To-Do item, along with other email meta-data.}
\vspace{-1.5em}
\label{email-eg}
\end{figure}

Smart To-Do takes a step further in task assistance and seeks to boost user productivity by automatically generating To-Do items from their email context. Text generation from emails, like creating To-Do items, is replete with complexities due to the diversity of conversations in email threads, heterogeneous structure of emails and various meta-deta involved. As opposed to prior works in text generation like news headlines, email subject lines and email conversation summarization, To-Do items are {\em action-focused}, requiring the identification of a specific task to be performed.%    since it must be relevant to the specific task outlined in the email. 

In this work, we introduce the task of automatically generating To-Do items from email context and meta-data to assist users with following up on their promised actions (also referred to as commitments in this work). Refer to Figure~\ref{email-eg} for an illustration. Given an email, its temporal context (i.e. thread), and associated meta-data like the name of the sender and recipient, we want to generate a short and succinct To-Do item for the task mentioned in the email.

\begin{figure}
\center
\includegraphics[width=0.4\textwidth]{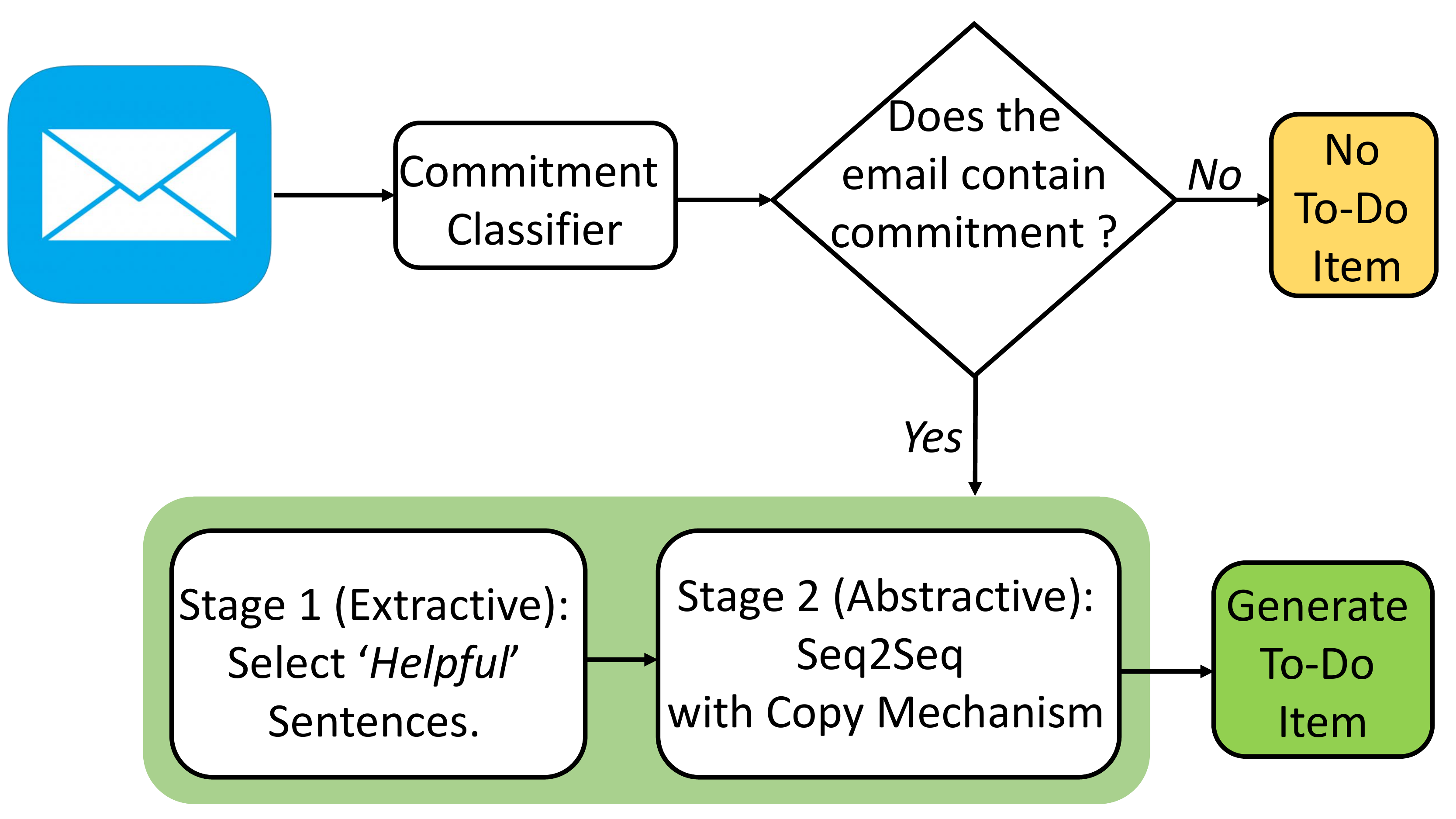} 
\vspace{-1em}
\caption{Smart To-Do flowchart: The email content is first scanned to detect any possible commitment sentence. If present, a To-Do item is generated using two-stage Smart To-Do framework.}
\label{fig:flowchart}
\vspace{-1em}
\end{figure}

This requires identifying the task sentence (also referred to as a {\textit{query}}), relevant sentences in the email that provide contextual information about the query along with the entities (e.g., people) associated with the task. We utilize existing work to identify the task sentence via a commitment classifier that detects action intents in the emails. Thereafter we use an unsupervised technique to extract key sentences in the email that are {\em helpful} in providing contextual information about the query. These pieces of information are further combined to generate the To-Do item using a sequence-to-sequence architecture with deep neural networks. Figure~\ref{fig:flowchart} shows a schematic diagram of the process. Since there is no existing work or dataset on this problem, our first step is to collect annotated data for this task.
%
%While generic summarization seeks to find information that is key to understanding the content of the entire email, task-specific summarization has to be more selective with regards to content extraction. 
%
%To ensure the appropriate selection of relevant sentences in the email thread conditioned on the query, we adapt a two-stage approach for text generation, similar to \cite{zhou2017selective, Chen2018FastAS, Zhang2019ThisEC}. From the current and previous email, we first rank sentences, in an unsupervised manner, to measure their relevance in writing the To-Do item. We then use the top $K$ ranked sentences in a sequence-to-sequence (Seq2Seq) architecture for To-Do item generation. 
%
%
Overall, our contributions can be summarized as follows:

\squishlist
\item We create a new dataset for To-Do item generation from emails containing action items based on the publicly available email corpus \textit{Avocado} \citep{avocado}. \footnote{We will release the code and  data (in accordance with LDC and Avocado policy) at \url{https://aka.ms/SmartToDo}. Email examples in this paper are similar to those in our dataset but are not reproducing text from the Avocado dataset.}
\item We develop a two-stage algorithm, based on unsupervised task-focused content selection and subsequent text generation combining contextual information and email meta-data.
\item We conduct experiments on this new dataset and show that our model performs at par with human judgments on multiple performance metrics.
\squishend

\section{Related Works}

Summarization of email threads has been the focus of multiple research works in the past \cite{rambow2004summarizing, carenini2007summarizing, dredze2008generating}. %Zhang2019ThisEC}. 
There has also been considerable research on identifying speech acts or tasks in emails \cite{carvalho2005speech,  lampert2010detecting, scerriLREC} and how it can be robustly adapted across diverse email corpora \citep{azarbonyad2019domain}. Recently, novel neural architectures have been explored for modeling action items in emails \cite{lin2018actionable} and identifying intents in email conversations \cite{Wang2019sigir}. However, there has been less focus on task-specific email summarization \citep{corston2004task}. 
The closest to our work is that of email subject line generation \citep{Zhang2019ThisEC}. But it focuses on a common email theme and uses a supervised approach for sentence selection, whereas our method relies on identifying the task-related context.

\section{Dataset Preparation}
We build upon the Avocado dataset~\citep{avocado}\footnote{Avocado is a more appropriate test bed than the Enron collection~\citep{klimt2004enron} since it contains additional meta-data and it entered the public domain via the cooperation and consent of the legal owner of the corpus.} containing an anonymized version of the Outlook mailbox for 279 employees with various meta-data and $938,035$ emails overall.

\subsection{Identifying Action Items in Emails} 

Emails contain various user intents including planning and scheduling meetings, requests for information, exchange of information, casual conversations, etc. \citep{Wang2019sigir}. For the purpose of this work, we first need to extract emails containing at least one sentence where the sender has promised to perform an action. It could be performing a task, providing some information, keeping others informed about a topic and so on. We use the term \textit{commitment} to refer to such intent in an email and the term \textit{commitment} sentence to refer to each sentence that has a \textit{commitment}.

\noindent \textbf{Commitment classifier:} A commitment classifier $\mathcal{C} : \mathcal{S} \mapsto [0, 1]$ takes as input an email sentence $\mathcal{S}$ and returns a probability of whether the sentence is a commitment or not. The classifier is built using labels from an annotation task with $3$ judges. The Cohen’s kappa value is $0.694$, depicting substantial agreement. The final label is obtained from the majority vote, generating a total of $9076$ instances (with $2586$ positive/commitment labels and $6490$ negative labels). The architecture is an RNN-based model with word embeddings and self-attention geared for binary classification with the input being the entire email context \citep{Wang2019sigir}. The classifier has a precision of $86\%$ and recall of $84\%$ on sentences in the Avocado corpus. 

\begin{savenotes}
\begin{table*}
\begin{tabular}{p{2.5cm}p{13cm}}
\toprule
{Ground-truth} & Update our quarterly sales in the head-office financial database.\\
{Annotation} & Update our quarterly sales in the database.\\
{Fluency} &   4 (Grammatically correct, follows structure of To-Do item.) \\
{Completeness} &  1 (Which \textit{database} ? Does not include additional details available from email context.)
\\
\midrule
{Ground-truth} & Test the server for load fault on Friday morning PST and let Bob know the result.\\
{Annotation} & Testing on server load fault on Friday morning PST and let Bob know the result. \\
{Fluency} &   2 (Grammatically incorrect; starts with `ing' verb and deviates from To-Do structure.) \\
{Completeness} &  4 (Explains the context and contains all keywords) \\
\bottomrule
\end{tabular}
\vspace*{-0.7em}
\caption{\label{score-eg} Snapshot of qualitative analysis of human annotations for fluency and completeness.}
\vspace*{-1.5em}
\label{tab:ann-analysis}
\end{table*}
\end{savenotes}

\subsection{To-Do Item Annotation}

%The Avocado corpus contains a total of $938,035$ emails (including duplicates) from $279$ accounts\footnote{We chose the Avocado corpus, as opposed to Enron, since Avocado is a larger and has close to double the number of user accounts in Enron}. Avocado reflects the enterprise setting effectively where emails containing commitments are more natural.
\noindent \textbf{Candidate emails:} We extracted $500k$ raw sentences from Avocado emails and passed them through the commitment classifier. We threshold the commitment classifier confidence to $0.9$ and obtained $29k$ potential candidates for To-Do items. Of these, a random subset of $12k$ instances were selected for annotation. 

\noindent \textbf{Annotation guideline:} 
For each candidate email $e_c$ and the previous email in the thread $e_p$ (if present), we obtained meta-data like `\textit{From}', `\textit{Sent-To}', `\textit{Subject}' and `\textit{Body}'. The commitment sentence in $e_c$ was highlighted and annotators were asked to write a To-Do item using all of the information in $e_c$ and $e_p$.  

We prepared a comprehensive guideline to help human annotators write To-Do Items containing the definition and structure of To-Do Items and commitment sentences, along with illustrative examples. Annotators were instructed to use words and phrases from the email context as closely as possible and introduce new vocabulary only when required. Each instance was annotated by $2$ judges.

\noindent \textbf{Analysis of human annotations:} %We now report some quantitative metrics obtained from the human annotations. 
We obtained a total of $9349$ email instances with To-Do items, each of which was annotated by two annotators. To-Do items have a median token length of $9$ and a mean length of $9.71$. %Given the email meta-data, the annotators were also asked whether the {\em subject} information was helpful in writing the To-Do Item. 
For $60.42\%$ of the candidate emails, both annotators agreed that the subject line was helpful in writing the To-Do Item.

To further analyze the annotation quality, we randomly sampled $100$ annotated To-Do items and asked a judge to rate them on (a) {\em fluency} (grammatical and spelling correctness), and (b) {\em completeness} (capturing all the action items in the email) on a $4$ point scale ($1$: Poor, $2$: Fair, $3$: Good, $4$: Excellent). Overall, we obtained a mean rating of 3.1 and 2.9 respectively for fluency and completeness. Table~\ref{tab:ann-analysis} shows a snapshot of the analysis.

%, denoting that it provided contextual information to understand the task. Oftentimes, the subject may be uninformative, but sentences in $e_c$  and $e_p$ prove useful in deciphering contextual information. There were also instances where a To-Do item could not be written due to missing information in the corpus or false positives from commitment classifier. 

%We also report the quantitative measure of overlap between the two judgments using BLEU-4 and Rouge metrics. 

%\begin{table}
%\centering
%\begin{tabular}{cccc}
%\hline \textbf{BLUE-4} & \textbf{Rouge-1} & \textbf{Rouge-2} & \textbf{Rouge-L}  \\ \hline
% 0.21 & 0.60  & 0.37 & 0.60\\
%\hline
%\end{tabular}
%\caption{\label{uhrs-metrics} Similarity of Human Judgements}
%\end{table}

\section{Smart To-Do : Two Stage Generation}

In this section, we describe our two-stage approach to generate To-Do items. In the first stage, we select sentences that are {\em helpful} in writing the To-Do item. Emails contain generic sentences such as salutations, thanks and casual conversations not relevant to the commitment task. The objective of the first stage is to select sentences containing informative concepts necessary to write the To-Do. %Moreover, if only this sentence in the email was available, we could understand the commitment made by the sender.

\subsection{Identifying Helpful Sentences for Commitment Task}

In the absence of reliable labels to extract helpful sentences in a supervised fashion, we resort to an unsupervised matching-based approach. Let the commitment sentence in the email be denoted as $\mathcal{H}$, and the rest of the sentences from the current email $e_c$ and previous email $e_p$ be denoted as $\{ s_1, s_2, \ldots s_d \}$. The unsupervised approach seeks to obtain a relevance score $\Omega(s_i)$ for each sentence. The top $K$ sentences with the highest scores will be selected as the extractive summary for the commitment sentence (also referred to as the query). 

\noindent \textbf{Enriched query context:} We first extract top $\tau$ maximum frequency tokens from all the sentences in the given email, the commitment and the subject (i.e., $\{ s_1, s_2, \ldots s_d \} \cup \mathcal{H} \cup \mathrm{Subject}$). Tokens are lemmatized and stop-words are removed. We set $\tau=10$ in our experiments. An enriched context for the query $\mathcal{E}$ is formed by concatenating the commitment sentence $\mathcal{H}$, subject and top $\tau$ tokens. 

\noindent \textbf{Relevance score computation:} Task-specific relevance score $\Omega$ for a sentence $s_i$ is obtained by inner product in the embedding space with the enriched context. Let $h(\cdot)$ be the function denoting the embedding of a sentence with $\Omega(s_i) = h(s_i)^Th(\mathcal{E})$.

Our objective is to find helpful sentences for the commitment given by semantic similarity between concepts in the enriched context and a target sentence. In case of a short or less informative query, the subject and topic of the email provide useful information via the enriched context. We experiment with three different embedding functions. 

(1) Term-frequency (Tf) -- The binarized term frequency vector is used to represent the sentence. 

(2) FastText Word Embeddings -- We trained FastText embeddings \citep{fasttext} of dimension $300$ on all sentences in the Avocado corpus. The embedding function $h(s_j)$  is given by taking the max (or mean) across the word-embedding dimension of all tokens in the sentence $s_j$.  %Although the training does not contain information combining email sentences together, it pertains to email structure and co-occurrence of words at sentence level. The embedding function $h(s_j)$ is then obtained by taking the max(or mean) across word-embedding dimensions of all tokens in the sentence $s_j$.

(3) Contextualized Word Embeddings -- We utilize recent advances in contextualized representations from pre-trained language models like BERT \citep{bert}. We use the second last layer of pre-trained BERT for sentence embeddings. 

We also fine-tuned BERT on the labeled dataset for commitment classifier. The dataset is first made balanced ($2586$ positive and $2586$ negative instances). Uncased BERT is trained for $5$ epochs for commitment classification, with the input being word-piece tokenized email sentences. This model is denoted as BERT (fine-tuned) in Table \ref{stage1-metrics}.
%A fine-tuned BERT on the commitment classification task was also used as another embedding option.
%A recent approach BERT \citep{bert} provides a pre-trained transformer network to obtain an embedding for a sentence. 

\noindent{\bf Evaluation of unsupervised approaches:} Retrieving at-least one helpful sentence is crucial to obtain contextual information for the To-Do item. Therefore, we evaluate our approaches based on the proportion of emails where at-least one helpful sentence is present in the top $K$ retrieved sentences.

We manually annotated $100$ email instances and labeled every sentence as \textit{helpful} or not based on (a) whether the sentence contains concepts appearing in the target To-Do item, and (b) whether the sentence helps to understand the task context. Inter-annotator agreement between $2$ judgments for this task has a Cohen Kappa score of $0.69$. This annotation task also demonstrates the importance of the previous email in a thread. Out of $100$ annotated instances, $44$ have a replied-to email of which $31$ contains a \textit{helpful} sentence in the replied-to email body ($70.4\%$). Table \ref{stage1-metrics} shows the performance of the various unsupervised extractive algorithms. FastText with max-pooling of embeddings performed the best and used in the subsequent generation stage.

%We use the following metric for evaluation that is in spirit similar to recall at a fixed $K$, albeit less strict. We measure whether out of $K$ top ranked sentences by the algorithm, at least one of them had the true annotated label as '\textit{Helpful}'. %The hope is that this sentence contains the necessary \textit{keywords} and along with the highlighted sentence, subject and address of Sent-To person will enable generation of To-Do item. 

\begin{table}
\centering
\small
\begin{tabular}{lcc}
\toprule
& \multicolumn{2}{c}{At-least One {Helpful}} \\
\textbf{Algorithm} & \textbf{@ K=2} & \textbf{@ K=3} \\ \midrule
Tf & 0.80 & 0.85\\
FastText (Mean) &  0.76 & 0.90 \\
FastText (Max) & \textbf{0.85} & \textbf{0.92} \\
BERT (Pre-trained) & 0.76 & 0.89 \\
BERT (Fine-tuned) &  0.80 & 0.89\\
\bottomrule
\end{tabular}
\vspace{-1em}
\caption{\label{stage1-metrics} Performance of unsupervised approaches in identifying {helpful} sentences for a given {query}.}
\vspace{-2em}
\end{table}

% \begin{table*}[ht!]
% \centering
% \small
% \begin{tabular}{ll}
% \toprule \textbf{Label} & \textbf{Generated Text}\\ \midrule
% GOLD  & Let john know about the training provide for booking resources.
% \\
% PRED  &  Let john know about booking resources.
% \\
% \midrule
% GOLD  & Put together the se plan and the overall day agenda of sales training.
%  \\
% PRED  &  Put together the draft agenda for sales training. \\
% \midrule
% GOLD  & Let elisabeth know about fedex package for hp.
% \\
% PRED  &  Let elisabeth know about fedex package for hp.
% \\\bottomrule
% GOLD  & Let alex know result.
% \\
% PRED  &  Let alex know about the license deal.
% \\
% \bottomrule
% \end{tabular}
% \vspace{-1em}
% \caption{\label{gen-eg} Generation examples ({\scriptsize GOLD: manual annotation, PRED: machine-generated}) with email context, meta-data in Appendix.}
% \vspace{-1em}
% \end{table*}

\begin{figure}
\center
\includegraphics[width=0.48\textwidth]{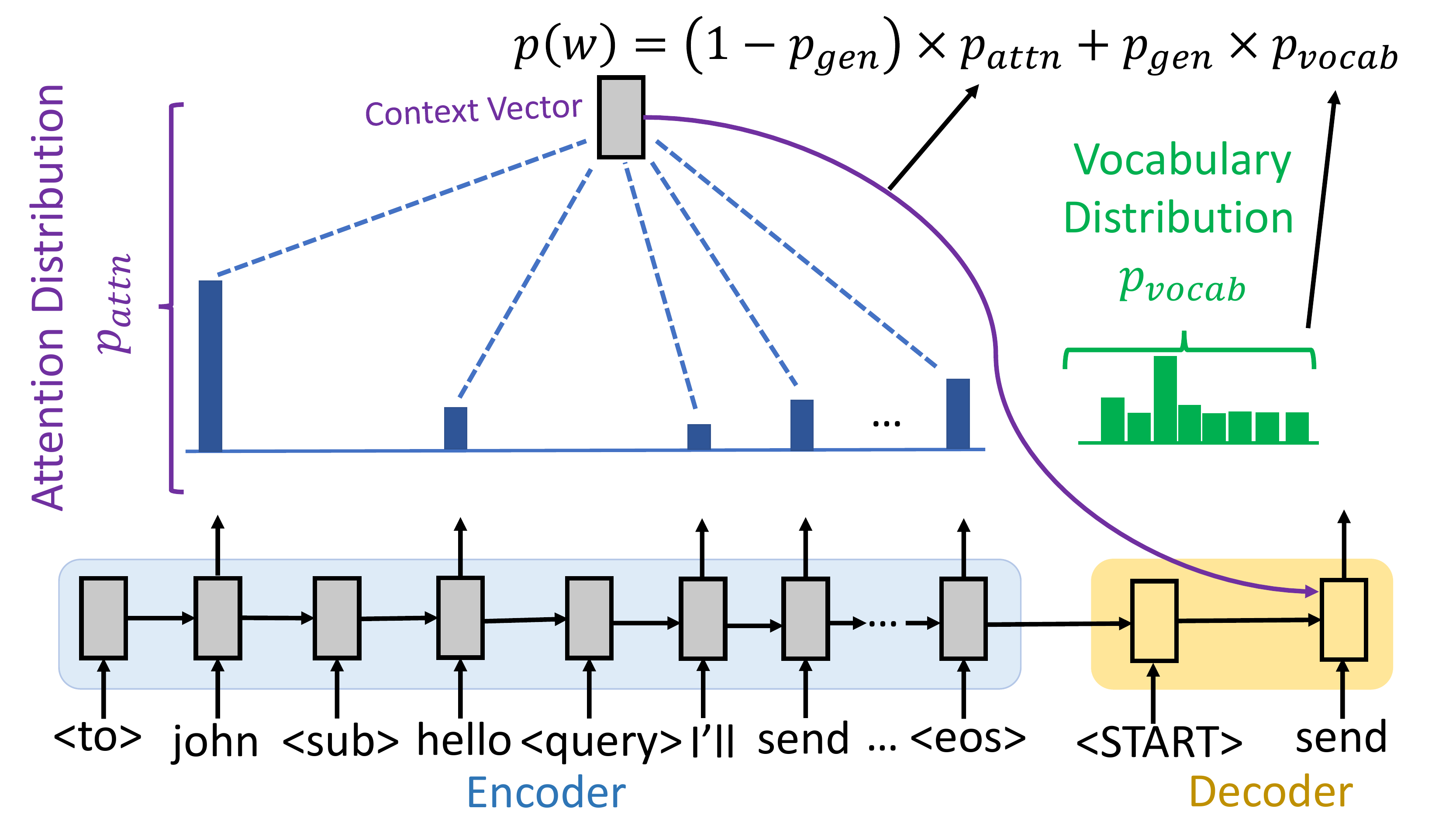} 
\caption{Seq2Seq with copy mechanism. Tokens involving named entities and task-specific keywords from the email are learned to copy in the To-Do item.}
\vspace{-1.5em}
\label{arcitecture}
\end{figure}

\subsection{To-Do Item Generation}

The generation phase of our approach can be formulated as sequence-to-sequence (Seq2Seq) learning with attention \cite{sutskever2014sequence, bahdanau2014neural}. It consists of two neural networks, an encoder and a decoder. The input to the encoder consists of concatenated tokens from different meta-data fields of the email like `{sent-to}', `{subject}', commitment sentence $\mathcal{H}$ and extracted sentences $\mathcal{I}$ separated by special markers. 
%In order to leverage important email meta-data, we use all tokens from different fields separated by markers as follows: 
%Since the To-Do item requires the  in first stage, we use all the tokens in these fields separated by the special symbols $<to>, <sub>, <high>$ and $<sent>$. 
For instance, the input to the encoder for the example in Figure \ref{email-eg} is given as:

{\small
\begin{lstlisting}[breaklines = True]
<to> alice <sub> hello ? <query> i will send it to you <sent> could you send me the sales report ? <eos>
\end{lstlisting}
}

We experiment with multiple versions of the generation model as follows:

\begin{savenotes}
\begin{table*}[t!]
\begin{tabular}{p{16cm}}
\toprule
\textit{From:} John Carter \qquad \textit{To:} Helena Watson; Daniel Craig; Rupert Grint \qquad \hspace*{\fill} \textit{Subject:} Thanks \\
\hspace{1cm} Thank you for helping me prepare the paper draft for ACL conference. Attached is the TeX file. Please feel free to make any changes to the revised version. I sent to my other collaborators already and am waiting for their suggestions. \hl{I'll keep you posted}. Thanks, John. \\
\midrule
GOLD: Keep Helena posted about paper draft for ACL conference. \\
\midrule
PRED: Keep Helana posted about ACL conference. \\
\bottomrule
\toprule
\textit{From:} Raymond Jiang\qquad \textit{To:}support@company.com \qquad \hspace*{\fill} \textit{Subject:} Bug 62 \\ 
\hspace{1cm} Hi, there is a periodic bug 62 appearing in my cellphone browser, whenever I choose to open the request. It might be a JavaScript issue on our side, but it would be nice if you take a look. Thanks, Ray. \\
\midrule
\textit{From:} Criag Johnson \qquad \textit{To:} Raymond Jiang \qquad \hspace*{\fill} \textit{Subject:} Bug 62 \\ 
Good Morning Ray, \hl{I shall take a look at it and get back to you}.\\
\midrule
GOLD: Take a look at Bug 62 and get back to Raymond. \\
\midrule
PRED: Take a look at periodic and get back to Raymond. \\
\bottomrule
\end{tabular}
\vspace*{-0.7em}
\caption{\label{gen-eg} Generation example ({GOLD: manual annotation, PRED: machine-generated}) with email context.}
\vspace{-1em}
\end{table*}
\end{savenotes}

\noindent \textbf{Vanilla Seq2Seq}: Input tokens $\{x_1, x_2, \ldots x_T\}$ are passed through a word-embedding layer and a single layer LSTM to obtain encoded representations $h_t = f(x_t, h_{t-1}) \, \forall \, t $ for the input. The decoder is another LSTM that makes use of the encoder state $h_t$ and prior decoder state $s_{t-1}$ to generate the target words at every timestep $t$. 
We consider Seq2Seq with attention mechanism where the decoder LSTM uses attention distribution $a_t$ over timesteps $t$ to focus on important hidden states to generate the context vector $h_t$. This is the first baseline in our work. 
\vspace{-0.5em}
\begin{align*}
e_{t,t'} &= v^T {tanh}(W_h \cdot h_t + W_s \cdot s_{t'} + b)  \\
a_{t,t'} &= {softmax}(e_{t,t'})\\
h_{t} &= \sum_{t'} a_{t,t'} \cdot h_{t'}
\end{align*}

\noindent \textbf{Seq2Seq with copy mechanism}: As the second model, we consider Seq2Seq with copy mechanism \citep{see2017get} to copy tokens from important email fields. Copying is pivotal for To-Do item generation since every task involves named entities in terms of the persons involved, specific times and dates when the task has to be accomplished and other task-specific details present in the email context. To understand the copy mechanism, consider the decoder input at each decoding step as $y_t$ and the context vector as $h_t$. The decoder at each timestep $t$ has the choice of generating the output word from the vocabulary $\mathcal{V}$ with probability $p_{\mathrm{gen}} = \phi(h_t, s_t, y_t)$, or with probability $1-p_{\mathrm{gen}}$ it can copy the word from the input context. To allow that, the vocabulary is extended as $\mathcal{V'} = \mathcal{V} \cup \{x_1, x_2, \ldots x_T\}$. The model is trained end-to-end to maximize the log-likelihood of target words (To-Do items) given the email context.

%email contents can vary drastically at test time from the trained corpus. The generated text needs to copy named entities, specific details and terms which are present in the input email tokens. Moreover, it needs to follow closely the structure of the commitment sentence. 

\begin{table}
\centering
%\small
\begin{tabularx}{0.45\textwidth}{lXXXX}
%\begin{tabular}{lllll}
\toprule \textbf{Algorithm} & {\bf \scriptsize BLEU-4} & {\bf \scriptsize Rouge-1} & {\bf \scriptsize Rouge-2}  & {\bf \scriptsize Rouge-L}\\ \midrule
Concatenate & 0.13  & 0.52  & 0.28  & 0.50 \\
Seq2Seq (vanilla) & 0.14  & 0.53 & 0.31 & 0.56 \\
Seq2Seq (copy) &  \textbf{0.23} & \textbf{0.60}   &  \textbf{0.41} &  \textbf{0.63}\\
Seq2Seq (BiFocal) & 0.18  &  0.56  & 0.34 & 0.58\\
\bottomrule
Human Judgment & 0.21  &  0.60  & 0.37 & 0.60\\
\bottomrule
\end{tabularx}
%\end{tabular}
%\vspace{-1em}
\caption{\label{overall-metrics} Comparison of various models for To-Do generation with BLEU and ROUGE (higher is better).}
%\vspace{-2em}
\vspace{-1em}
\end{table}

\noindent \textbf{Seq2Seq BiFocal}: As a third model, we experimented with query-focused attention having two encoders -- one containing only tokens of the query and the other containing rest of the input context. We use a bifocal copy mechanism that can copy tokens from either of the encoders. We refer the reader to the Appendix for more details about training and hyper-parameters used in our models.

\section{Experimental Results}

We trained the above neural networks for To-Do item generation on our annotated dataset. Of the $9349$ email instances with To-Do items, we used $7349$ for training and $1000$ each for validation and testing. For each instance, we chose the annotation with fewer tokens as ground-truth reference. 
 
The median token length of the encoder input is $43$ (including the {helpful} sentence). Table \ref{overall-metrics} shows the performance comparison of various models. We report BLEU-4 \citep{papineni2002bleu} and the F1-scores for Rouge-1, Rouge-2 and Rouge-L~\citep{lin2004rouge}. %We include as baseline vanilla Seq2Seq with attention, where no copy mechanism is present. 
We also report the human performance for this task in terms of the above metrics computed between annotations from the two judges.

A trivial baseline -- which concatenates tokens from the `{sent-to}' and `{subject}' fields and the commitment sentence -- is included for comparison. 

The best performance is obtained with Seq2Seq using copying mechanism. We observe our model to perform at par with human performance for writing To-Do items. Table \ref{gen-eg} shows some examples of To-Do item generation from our best model. %We found the query-focused bifocal copy mechanism to be slightly worse than using a single copy-mechanism resulting from the ambiguity on whether to copy from the query or the context at a given timestep. 

\section{Conclusions}

In this work, we study the problem of automatic To-Do item generation from email context and meta-data to provide smart contextual assistance in email applications. To this end, we introduce a new task and dataset for action-focused text intelligence. We design a two stage framework with deep neural networks for task-focused text generation. 

There are several directions for future work including better architecture design for utilizing structured meta-data and replacing the two-stage framework with a multi-task generation model that can jointly identify helpful context for the task and perform corresponding text generation.

%Future work will focus on combining the two stages through joint training. %two-stage pipeline to solve the problem. %Future work will explore a joint training of the two stages to further improve performance.

\bibliography{anthology,acl2020}
\bibliographystyle{acl_natbib}

\appendix

\section{Appendix}

\subsection{Hyper-parameters}

We now provide the hyper-parameters and training details for ease of reproducibility of our results. The encoder-decode architecture consists of LSTM units. The word embedding look-up matrix is initialized using Glove embeddings and then trained jointly to adapt to the structure of the problem. We found this step crucial for improved performance. Using random initialization or static Glove embeddings degraded performance. 

We also experimented with using either a shared or a separate vocabulary for the encoder and decoder. A token was included in the vocabulary if it occurred at least $2$ times in the training input/target. Separate vocabulary for source and target had better performance. Typically, source vocabulary had higher number of tokens than target. A shared dictionary led to increased number of parameters in the decoder and to subsequent over-fitting. The validation data was used for early stopping. The patience was decreased whenever either the validation token accuracy or perplexity failed to improve. We used the OpenNMT framework in PyTorch for all our Seq2Seq experiments. 

Table \ref{seq2seq-param} lists the hyper-parameters of the best performing model.

\begin{table}[h!]
\begin{center}
\begin{tabular}{ll}
\multicolumn{1}{c}{\bf Hyper-parameter}  &\multicolumn{1}{c}{\bf Value} \\
\hline \\
Rnn-type & LSTM \\ 
Rnn-size &  256 \\
\# Layers & 1 \\ 
Word-embedding & 100 \\
Embedding init. & Glove \\
Batch size & 64 \\
Optimizer & Adagrad \\
Learning rate & 0.15 \\
Adagrad accumulator init. & 0.1 \\
Max. Gradient norm & 2.0 \\
Dropout & 0.5 \\
Attention dropout & 0.5 \\
Tokenizer & spacy \\
Vocabulary & Separate \\
Early Stopping (Patience) & 5\\
Beam width & 5 \\
\end{tabular}
\end{center}
\caption{Seq2Seq with copy mechanism : Hyper-parameters for the best model.}
\label{seq2seq-param}
\end{table}

\subsection{Illustrative Examples}

In this Section, we provide further examples of the email threads along with the highlighted commitment sentence. Note that some of the emails have previous thread email present, and some do not have it. For each of these examples, we also provide the To-Do item written by the human judge (denoted as GOLD) and that predicted by our best model (denoted as PRED). As in the main text, the sentences have been paraphrased and names changed due to the data sensitivity of Avocado.

\begin{table*}
\begin{tabular}{p{16cm}}
\Xhline{2.0pt}
\textit{From:} Beverly Evans \qquad \textit{To:} Carlos Simmons \qquad \hspace*{\fill} \textit{Subject:} Amazon.com update \\ 
Carlos, \\
\hspace{2cm} I came to know today from John Carter than we received a PHP script that is not decoding the correct database. Can you check with them why they sent us the eCommerce PHP code when the loss of functionality was not out fault? I have registered the error log in the eCommerce section because the staff scientist from Amazon mentioned it in his email. He also said they have not been able to resolve the issue and surprisingly did not mention who we should contact next. (This email exchange was about a week ago when I had handed them the cloud expenditures.) Also, we need to generate a PHP example to replicate the error. Could you update me if the team is working on it?\\
Thanks, Beverly \\
\hline
\textit{From:} Carlos Simmons \qquad \textit{To:} Beverly Evans \qquad \hspace*{\fill} \textit{Subject:} Amazon.com update \\ 
\hspace{2cm} The PHP they shared with us is an example. eCommerce is not what they want us to resolve. I feel we should wait until their engineers test all possibilities. Joseph informed us that they need to test the database more carefully and figure out which PHP code to send to us and whether they want our feedback on the database. I am not sure why they sent me a 'relevant PHP example' - I thought there was the only file they sent us yesterday. \hl{I will forward that to you and Renata}. \\
\hline
GOLD: Forward PHP example to Beverly and Renata. \\
\hline
PRED: Forward eCommerce PHP to Beverly. \\
\Xhline{2.0pt}
\end{tabular}
\caption{\label{illustrative-eg1} Illustrative Example 1}
\end{table*}

\begin{table*}
\begin{tabular}{p{16cm}}
\Xhline{2.0pt}
\textit{From:} Kirstin Barnes  \qquad \textit{To:} Nannie Jacobs \qquad \hspace*{\fill} \textit{Subject:} Ready for Product Launch \\
Nannie, \\ 
\hspace{2cm} I am ready for the product launch. I need to include some of the enhancements in the presentation. \hl{I'll submit what is already completed and then do the remaining after the meeting.}. \\
Kirstin Barnes  \\
Product Engineer AvocadoIT, Inc. \\
\hline
GOLD: Submit presentation with product enhancements. \\
\hline
PRED: Submit the enhancements for product launch. \\
\Xhline{2.0pt}
\end{tabular}
\caption{\label{illustrative-eg2} Illustrative Example 2}
\end{table*}

\begin{table*}
\begin{tabular}{p{16cm}}
\Xhline{2.0pt}
\textit{From:} Rishabh Iyer \qquad \textit{To:} R\&D \qquad \hspace*{\fill} \textit{Subject:} Software not ready yet for deployment \\
Hello, \\ 
\hspace{2cm} Unlike our plan last month, the software is still not ready for deployment. The team put together some errors last week. We must plan to make it available latest by next week. \hl{I will keep you posted}. \\
Thanks, Rishabh Iyer. \\
Software Engineer AvocadoIT Inc. \\
\hline
GOLD: Keep r\&d posted about deployment of software. \\
\hline
PRED: Keep r\&d posted about deployment. \\
\Xhline{2.0pt}
\end{tabular}
\caption{\label{illustrative-eg3} Illustrative Example 3}
\end{table*}

\begin{table*}
\begin{tabular}{p{16cm}}
\Xhline{2.0pt}
\textit{From:} Justine Sparrow \qquad \textit{To:} Roma Patterson \qquad \hspace*{\fill} \textit{Subject:} 24x7 Helpline \\
Roma, \\ 
\hspace{2cm} I will bring this up in the Staff meeing today. \hl{I'll let you know the outcome}. Could you confirm if this is for a license agreement or a shared solution  ? \\
Thanks, Justine. \\
\hline
GOLD: Let Roma know result. \\
\hline
PRED: Let Roma know about the license agreement. \\
\Xhline{2.0pt}
\end{tabular}
\caption{\label{illustrative-eg4} Illustrative Example 4}
\end{table*}

% \begin{table*}
% \begin{tabular}{p{16cm}}
% \Xhline{2.0pt}
% \textit{From:} Joseph Raymond \qquad \textit{To:}support@avocadoit.com \qquad \hspace*{\fill} \textit{Subject:} Err 4 \\ 
% Hi, \\
% \hspace{2cm} I am getting a periodic err 4 from the quotes request using sprint pcs touchpoint phone. It may be an xml err on our side, but can you take a look. \\
% Thanks, Joe Raymond. \\
% \hline
% \textit{From:} Brian Robinson \qquad \textit{To:} Joseph Raymond \qquad \hspace*{\fill} \textit{Subject:} Err 4 \\ 
% Good Morning Joe, \hl{I'll take a look at it and get back to you}.\\
% \hline
% GOLD: Take a look at Err 4 and get back to Joseph. \\
% \hline
% PRED: Take a look at periodic and get back to joseph \\
% \Xhline{2.0pt}
% \end{tabular}
% \caption{\label{illustrative-eg5} Illustrative Example 5}
% \end{table*}

\begin{table*}
\begin{tabular}{p{16cm}}
\Xhline{2.0pt}
\textit{From:} Rebecca Anderson \qquad \textit{To:} Julia Roberts \qquad \hspace*{\fill} \textit{Subject:} Run a bash script while synchronize \\ 
Julia, \\
\hspace{2cm} When synchronizing is done, we want to run a bash script to delete old records on the machine and remove all activity logs. How can I do this ? What is the way to perform this operation ? Also, in the bash script, is there a way to sort the dates so that we can identify older activities ? \\
Thanks, Rebecca. \\
\hline
\textit{From:} Julia Roberts \qquad \textit{To:} Rebecca Anderson \qquad \hspace*{\fill} \textit{Subject:} Run a bash script while synchronize \\ 
Rebecca, \\
We had exactly the same feature to delete activities which you mentioned in our previous release. But we no longer have that in the new version due to resource constraints. \hl{I will take to John to review this again}. \\
Thanks, Julia. \\
\hline
GOLD: Talk to John to review bash script again. \\
\hline
PRED: Talk to John to review the activities. \\
\Xhline{2.0pt}
\end{tabular}
\caption{\label{illustrative-eg6} Illustrative Example 6}
\end{table*}

\begin{table*}
\begin{tabular}{p{16cm}}
\Xhline{2.0pt}
\textit{From:} Ramesh Paul \qquad \textit{To:} Gopal Majumdar \qquad \hspace*{\fill} \textit{Subject:} Updates List for 3/11 \\
\hspace{2cm} Here's the update for this week. 1. The R\&D team is working on a presentation for the knowledge tranfer for v5. It should be ready within next two weeks. 2. I have received their email, but need to review the ppt. 3. Did you want to know more about the new cloud feature for automatic version management ? Or was it a different feature ? 4. I am constantly working on this. 5. Didn't we discuss this point in our last email ? 6. We are making similar tests in the desktop for v5 before migrating to the cloud. We first have to make sure things work well for the desktop. \hl{I will send you more details soon}. Did you get a chance to update your blog with information about these new features ? \\
Thanks, Ramesh. \\
\hline
GOLD: Send Gopal more details about tests in the desktop for v5. \\
\hline
PRED: Send Gopal more details on presentation for the knowledge transfer. \\
\Xhline{2.0pt}
\end{tabular}
\caption{\label{illustrative-eg8} Illustrative Example 8}
\end{table*}

\begin{table*}
\begin{tabular}{p{16cm}}
\Xhline{2.0pt}
\textit{From:} Lori Howard \qquad \textit{To:} Karen James; Bruce Thomas; Steve Perry \qquad \hspace*{\fill} \textit{Subject:} Room reservations \\
Team, \\
\hspace{2cm} This needs to be done through a formal training session, but as of now let me point out some crucial points about room reservations. 1. In case you allocate a room for general meetings and administrative work, then make sure you book it for that month, but not for long periods of time. (Karen, can you check with Renata whether this is fulfilled for our meetings next week?) 2. In case of clients who do not need the entire month, make sure to reserve only for the particular month. If it exceeds that time, the system will authomatically resolve it and reserve it for next month. 3. For room reservation, either enter the number of hours required or the \% of month, but not both. I would prefer precise hours. \hl{I will inform you when we can provide training, perhaps we can next week}. \\
Thanks, Lori. \\
\hline
GOLD: Let Karen know about the training provide for room reservations. \\
\hline
PRED: Let Karen know about room reservations. \\
\Xhline{2.0pt}
\end{tabular}
\caption{\label{illustrative-eg9} Illustrative Example 9}
\end{table*}

\begin{table*}
\begin{tabular}{p{16cm}}
\Xhline{2.0pt}
\textit{From:} Matthew White \qquad \textit{To:} Frank; Paul; Dennis \hspace*{\fill} \textit{Subject:} Draft Agenda for Software Training \\
Dear All, \\
\hspace{2cm} As discussed before, we have finally come to a concrete plan. I have attached the draft for your review. Please go over it and let me know asap your suggestions so that I can send them to the organizers. Please check the agenda and the names of trainees. \hl{I'll put together the Training plan and the overall 5-day agenda as soon as I can.}\\
Matthew. \\
\hline
GOLD: Put together the training plan and the overall day agenda of software training. \\
\hline
PRED: Put together the draft agenda for software training. \\
\Xhline{2.0pt}
\end{tabular}
\caption{\label{illustrative-eg10} Illustrative Example 10}
\end{table*}

\begin{table*}
\begin{tabular}{p{16cm}}
\Xhline{2.0pt}
\textit{From:} Diana Wilson \qquad \textit{To:} Alba Deacon \qquad \hspace*{\fill} \textit{Subject:} DHL package from IBM \\ 
Alba, \\
\hspace{2cm} I was able to track the package and as per the website it was in Sao Luis, Brazil at noon. I am not sure where it is, but it is Brazil so ... Send me an update if you receive it from them. 

I just tracked the package and as of 10:00am today it was in Toluca, Mexico. Where that is I have no idea but it is in Mexico so ... Let me know if you hear from them when they receive it. \\
Thanks. Diana Wilson. \\
\hline
\textit{From:} Alba Deacon \qquad \textit{To:} Diana Wilson \qquad \hspace*{\fill} \textit{Subject:} DHL package from IBM \\ 
Thanks Diana. \hl{If I hear anything I'll let you know.}.\\
~ Alba. \\
\hline
GOLD: Let Diana know about DHL package from IBM. \\
\hline
PRED: Let Diana know about DHL package from IBM. \\
\Xhline{2.0pt}
\end{tabular}
\caption{\label{illustrative-eg11} Illustrative Example 11}
\end{table*}

% \begin{figure*}
% \center
% \includegraphics[width=0.8\textwidth]{Images/HitApp.pdf} 
% \caption{HitApp Environment}
% \label{flowchart}
% \end{figure*}

% \end{document}

%\section{Supplementary - Hyperparameter tuning}

\end{document}